\pdfoutput=1

\documentclass[11pt]{article}

\usepackage{acl}

\usepackage{pifont}

\usepackage{times}
\usepackage{latexsym}
\usepackage{amsmath}
\usepackage{amsmath,amssymb}

\usepackage[T1]{fontenc}

\usepackage[utf8]{inputenc}

\usepackage{microtype}

\usepackage{inconsolata}

\usepackage{graphicx}
\usepackage{pifont}
\usepackage{caption} 
\usepackage{float}
\usepackage{multirow}
\usepackage{tabularx,colortbl}
\usepackage{subcaption}

\title{Efficient Architectures for High Resolution Vision-Language Models}

\author{Miguel Carvalho \ ~~~~~~~~~ \ Bruno Martins\\
INESC-ID and Instituto Superior Técnico, University of Lisbon \\ 
\texttt{\{miguelcarvalho00,bruno.g.martins\}@tecnico.ulisboa.pt}}

\begin{document}
\maketitle
\begin{abstract}
Vision-Language Models (VLMs) have recently experienced significant advancements. However, challenges persist in the accurate recognition of fine details within high resolution images, which limits performance in multiple tasks. This work introduces Pheye, a novel architecture that efficiently processes high-resolution images while training fewer parameters than similarly sized VLMs. Notably, Pheye achieves a high efficiency while maintaining strong performance, particularly in tasks that demand fine-grained image understanding and/or the handling of scene-text.
\end{abstract}

\section{Introduction}

The integration of visual capabilities as extensions to Large Language Models (LLMs) has led to the emergence of large Vision-Language Models (VLMs), currently excelling in tasks like image captioning and visual question answering. Notable examples include Flamingo~\citep{alayrac2022flamingo} or Monkey~\citep{li2023monkey}, in this last case bringing improvements to tasks that involve scene-text by processing higher-resolution images. LLaVA-NeXT~\citep{liu2024llavanext} followed Monkey’s lead with similar enhancements, but at the cost of a quadratic increase in the computational complexity of the language model, as it skipped a resampler module that compresses large sequence lengths, from the encoder, into a fixed size of query vectors.

Recent focus has shifted towards smaller VLMs that can run on hardware-constrained devices. Models like MoE-LLaVA~\citep{lin2024moe} or moondream2~\citep{moondream2} achieve impressive performance with a fraction of the parameters of their predecessors. This work further advances small VLMs with Pheye, i.e. a family of compact models that can process high-resolution images with fewer parameters and computational demands, expanding VLM applications to resource-limited environments where understanding fine details is crucial.

Specifically, Pheye employs a frozen instruction-tuned language model~\citep{li2023textbooks} in conjunction with a frozen pre-trained CLIP~\citep{radford2021learning} vision model, which are linked by dense cross-attention layers inserted before the language model's layers. To process high-resolution images, we use two sets of LoRA~\citep{hu2021lora} adapters in the vision encoder, one for the global image and another for local high-resolution patches. 

\begin{figure}[t!]
        \centering
        \includegraphics[scale=.28]{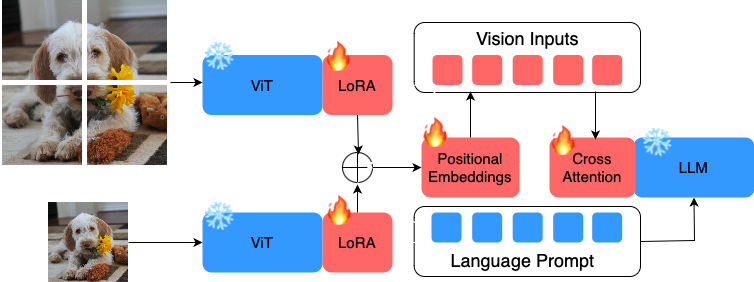}
        \vspace{-0.25cm}
        \caption{Overview on the proposed architecture, where input images are split into regular non-overlapping patches that match the input resolution of a pre-trained ViT. Two sets of LoRA adapters are respectively used to adjust the ViT to both global and local sub-images, and a frozen LLM is conditioned on the concatenated vision representations through dense cross-attention layers.}
        \vspace{-0.5cm}
        \label{fig:architecture}
\end{figure}

Notably, Pheye is competitive with similarly sized models, particularly in tasks involving scene-text, such as TextVQA~\citep{singh2019towards}. Pheye requires only a fraction of the training parameters, being more efficient in connecting the vision and language modalities, and in processing input images at higher resolutions. The source code associated to the experiments reported on this paper, as well as the trained models, are available from a public GitHub repository\footnote{\url{https://github.com/miguelscarv/pheye}}.

\section{An Efficient High-Resolution VLM}

This section presents the proposed method for building high-resolution efficient VLMs, starting with architectural design choices, and then analysing the model's computational complexity.

\subsection{The Proposed Architecture}
\label{sec: crossdense}

The Pheye architecture is illustrated in Figure~\ref{fig:architecture}. It employs a frozen instruction-tuned language model and a vision encoder that adapts a pre-trained CLIP model, linking the two components with dense cross-attention layers inserted before the language model's layers. The use of cross-attention and the design of the vision encoder were informed by preliminary experiments described in Appendix~\ref{sec:appendix}. 
    
The vision encoder consists of a ViT with two sets of LoRA adapters, one for encoding global images and another for local high-resolution patches, as outlined in Appendix~\ref{sec: vision encoder}. To better distinguish between global and local patch embeddings and improve convergence, we introduce two LayerNorm~\citep{ba2016layer} layers after the ViT, and before adding learned positional embeddings. These layers are applied separately, respectively processing the global and local patch embeddings.

\begin{figure}[t!]
        \centering
        \includegraphics[scale=.25]{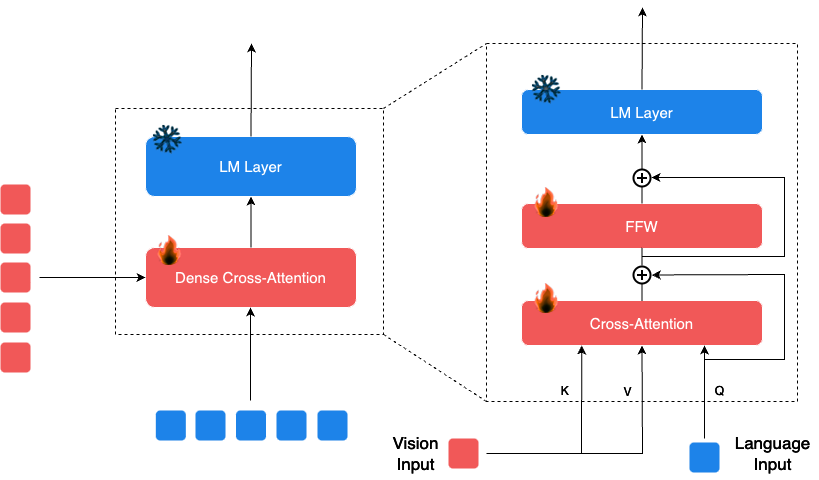}
        \vspace{-0.25cm}
        \caption{An illustration for dense cross-attention layers. To condition the language model on visual inputs, we add new cross-attention layers between existing pre-trained and frozen language model layers. The keys and values for these layers are derived from vision features, while the queries come from language inputs. These layers are followed by dense feed-forward layers. The output matrices of both of these modules are initialized with values close to zero to maintain the integrity of the language model at initialization.}
        \vspace{-0.5cm}
        \label{fig:crossdense}
\end{figure}

Given the computational efficiency, strong task performance, and reduced number of trainable parameters, we use cross-attention to combine modalities while keeping the language model frozen. Inspired by the Flamingo architecture~\citep{alayrac2022flamingo}, we replace vanilla cross-attention modules with dense cross-attention modules, inserting them at regular intervals before the decoder layers, as shown in Figure~\ref{fig:crossdense}. This design was motivated Flamingo's demonstration that gated cross-attention outperforms vanilla cross-attention even when parameter counts are equal. However, we deviate from Flamingo's gated cross-attention layers, as early experiments showed that the gating mechanism hindered convergence. To preserve the language model's initial integrity, we initialize the output matrices of the dense cross-attention modules with values sampled from a normal distribution with a mean of zero and a variance approaching zero. This way, the cross-attention layers have a minor influence during the initial training epochs. 

\vspace{-0.10cm}
\subsection{Analysis of the Computational Complexity}
\vspace{-0.05cm}

This section analyzes the computational complexity of the proposed methods, without accounting for the LoRA adapters in the vision encoder, by calculating the number of multiplications that are involved in all linear layers and attention operations, separately considering the vision encoder and the language model with dense cross-attention layers. The analysis assumes a sequential order of operations in all linear layers and attention mechanisms, although implementation optimizations can be latter used. We compare our method with the most widely used approach of using a higher-resolution ViT and a LLaVA-style architecture.

For reference, the cost of a matrix multiplication $C$ between matrices $N \in\mathbb{R}^{n \times d}$ and $W \in\mathbb{R}^{d \times o}$ would be $C = (n \times o) \times d$, where $N$ represents a matrix of $n$ rows with dimensionality $d$, and $W$ represents a weight matrix with an input dimensionality of $d$ and an output dimensionality of $o$. 

\paragraph{Vision Encoder.} The computational complexity of a ViT depends on the number of input tokens $N$, corresponding to the number of image patches plus one for the \texttt{[CLS]} token, and the model dimensionality $D$. The number of multiplications for a single Transformer layer can be expressed as:
\begin{equation} \label{vitformula} \small
\vspace{-0.01cm}
    \mathrm{\bold{T}_{ViT}} = 4ND^2 + DN^2 + 8ND^2.
\vspace{-0.05cm}
\end{equation}
This complexity breaks down into three components, namely (i) the multiplications in the $W_Q$, $W_K$, $W_V$ and $W_O$ matrices, (ii) the attention mechanism, and (iii) a feed-forward module with two layers, where the intermediate dimensionality is four times the model dimensionality.

In the case of our strategy, instead of computing self-attention across all the high-resolution input tokens, the image is broken down into $P$ sub-images of equal size and lower-resolution, and the number of multiplications can be expressed by the following equation, where $N'$ is now reflecting the number of patches per sub-image, plus one \texttt{[CLS]} token also per sub-image (i.e., each sub-image involves a total of $N'= \frac{(N-1)}{P-1} + 1$ tokens).
\begin{equation} \label{vitformula2} \small
    \mathrm{\bold{T}_{Pheye}} = (4N'D^2 + DN'^2 + 8N'D^2) \times P.
\end{equation}

We can compare the efficiency of our vision encoder against a ViT that operates at a resolution of 672$\times$672 pixels with a patch size of 14 pixels, resulting in $N=2305$ input tokens. Our method would process 10 images, where 1 is global and 9 are local sub-images, at a resolution of 224$\times$224 pixels with a patch size of 14, yielding $N'=2570$ tokens. With $D = 1280$, our method is approximately 1.02 as efficient as the alternative. While this corresponds to little to no improvement, our method does offer a significant advantage: it eliminates the need to fine-tune the underlying ViT, allowing us to focus solely on training the LoRA parameters to increase the input resolution.

\paragraph{Language Model.} For the language model, we analyze the computational complexity of a LLaVA-style architecture, against that of our method. The complexity of the LLaVA style architecture per layer is given by Equation~\ref{llavacomplexity} and our method has an average complexity per layer given by Equation~\ref{pheyecomplexity}, where the fractional term refers to the average complexity introduced by dense cross-attention layers. Specifically, the aforementioned term has 4 components in the numerator: (i) the multiplications in the $W_K$ and $W_V$ matrices, (ii) the multiplications in $W_Q$ and $W_O$, (iii) the attention mechanism, and (iv) the feed-forward module. In both formulas, $N_T$ represents the number of text tokens, $N_I$ represents the number of image tokens, $D$ and $D_{ViT}$ correspond to the dimensionality of the language model and the ViT, respectively, and $I$ denotes the interval at which dense cross-attention layers are inserted in the language model.
\begin{equation}
    \label{llavacomplexity}
    \small
    \begin{split}
    \mathrm{\bold{T}_{LLaVA}} = & 4(N_T+N_I)D^2 + D(N_T+N_I)^2 + \\
    & 8(N_T+N_I)D^2.
    \end{split}
\end{equation}
\begin{equation}
    \label{pheyecomplexity} 
    \small
    \begin{split}
        & \mathrm{\bold{T}_{Pheye}} = 4N_TD^2 + DN_T^2 + 8N_TD^2 + \\
        & \frac{2N_ID_{ViT}D + 2N_TD^2 + DN_TN_I + 8N_TD^2}{I}.
    \end{split}
\end{equation}

For assessing the efficiency gain of our language model against a LLaVA-style architecture, assume $N_T = 65$, which is the typical prompt length for captioning, $N_I = 2305$, corresponding to the number of vision tokens output by a ViT with an input resolution of 672x672 pixels, $I = 2$, $D = 2048$ and $D_{ViT} = 1280$. Our method is, in this case, approximately 12.1 times more efficient than its LLaVA-style counterpart. Furthermore, if we increase the interval to $I = 4$, our method becomes approximately 18.5 times more efficient than the corresponding LLaVA-style architecture.

\section{Main Experimental Evaluation}

This section begins by outlining the Pheye training setup. It then presents the results achieved by our models on academic task-oriented datasets.

\subsection{Experimental Setup}

Our experiments aimed to assess the effects of increasing the input image resolution, and to examine the impact of augmenting the frequency of dense cross-attention layers in the language model. To do this, we compare four model settings, varying the image input resolution between 448$\times$448 pixels and 672$\times$672 pixels, and adjusting the dense cross-attention interval to every 4 or 2 decoder layers. 

In all four settings, we used a Phi 1.5 language model~\citep{li2023textbooks}, finetunned on the SlimOrca~\citep{SlimOrca} instruction dataset. The ViT is initialized from a CLIP-ViT-H-14 model finetunned on DFN-5B~\citep{fang2023data}. 

The models were trained in separate stages, similarly to MoE-LLaVA \citep{lin2024moe}, using a cross-entropy loss over the output tokens. The different stages are described next, while Appendix \ref{sec:datamix} summarizes the datasets used in Stage III.

\begin{table*}[t!]
    \centering
    \small
    \begin{tabular}{p{1.3in}cccccccccccc}
              & \multicolumn{3}{c}{Model Size} &  & \multicolumn{5}{c}{Evaluation Results} \\
              \cline{2-4} \cline{6-10} \\[-1.5ex]
        Model & Res. & Trained & Data & & VQAv2 & NoCaps & TextVQA & TextCaps & DOCCI \\
        \hline & \\[-1.5ex]
        BLIP-2 & 224 & 187M & 129M & & 63.0 & 104.5 & 43.1$^*$ & - & - \\
        InstructBLIP-2 & 224 & 187M & 130M & & - & 119.9 & 46.6$^*$ & 82.4$^{\dag}$ & 5.7$^{\dag}$ \\
        MobileVLM 1.7B & 336 & 1.4B & 3.9M & &  - & - & 41.5 & - & - \\
        MobileVLM V2 1.7B & 336 & 1.4B & 6.3M & &  - & 90.0$^{\dag}$ & 52.1$^*$ & 48.5$^{\dag}$ & 4.5$^{\dag}$ \\
        MoE-LLaVA-1.8B×4 & 336 & 2.8B & 6.6M & & 76.2 & - & 48.0$^*$ & - & - \\
        MoE-LLaVA-2.7B×4 & 336 & 5.0B & 6.6M & & 77.6 & - & 51.4$^*$ & - & - \\
        moondream1 & 384 & 1.86B & 3.9M & & 74.7 & - & 35.6 & - & - \\
        moondream2 & 384 & 1.86B & - & & 77.7 & 92.5$^{\dag}$ & 49.7 & 120.2$^{\dag}$ & 0.2$^{\dag}$ \\
        \hline & \\[-1.5ex]
        Pheye-x4 & 448 & 295M & 2.9M & & 75.2 & 110.3 & 45.9 & 106.4 & 5.7 \\
        Pheye-x4 & 672 & 295M & 2.9M & & 75.5 & 110.8 & 49.2 & 111.9 & 5.4 \\
        Pheye-x2 & 448 & 578M & 2.9M & & 76.0 & 111.8 & 47.3 & 108.9 & 5.6 \\
        Pheye-x2 & 672 & 578M & 2.9M & & 76.4 & 110.5 & 50.5 & 115.9 & 5.9 \\
        \hline & \\[-1.5ex]
        Pheye-x2 {\it \tiny (upscaled low-res inputs)} & 672 & 578M & 2.9M & & 76.3 & 112.0 & 44.3 & 105.7 & 5.7 \\
        \hline & \\[-1.5ex]
    \end{tabular}
\caption{Results on different academic task-oriented datasets. "Res.", "Trained" and "Data" represent the input image resolution, the number of trainable parameters, and the number of training instruction-response pairs, respectively. Both BLIP-2~\citep{li2023blip} and InstructBLIP-2~\citep{dai2024instructblip} refer to the FlanT5\textsubscript{XL} \citep{chung2024scaling} variants, while moondream2 refers to the 2024-04-02 model version. The symbol \textsuperscript{*} denotes evaluations made with OCR tokens in the instruction, while \textsuperscript{\dag} refers to our own evaluation of the models, using the prompt \texttt{"Provide a one-sentence caption for the provided image"}, or instead the prompt \texttt{"Generate a highly detailed description for the provided image using multiple sentences"} for the case of DOCCI~\citep{OnoeDocci2024}, without OCR tokens in the instruction. Pheye-xN denotes a Pheye model with dense cross-attention layers inserted every $N$ layers of the Phi 1.5 model. VQAv2 refers to the test-dev split, NoCaps~\citep{agrawal2019nocaps} and TextVQA refer to the validation splits, and TextCaps and DOCCI refer to the test splits of the corresponding datasets. VQAv2 and TextVQA report VQA accuracy, while NoCaps, TextCaps, and DOCCI report CIDEr.}
\label{tab:main_results}
\vspace{-0.20cm}
\end{table*}

\vspace{-0.175cm}
\paragraph{$\bullet$ Stage I.} We initially aimed for a model that can effectively describe images, including their finer details. We used ShareGPT4V-PT~\citep{chen2023sharegpt4v}, featuring 1,246K images with detailed descriptions, of which 570K are high resolution images from SAM~\citep{kirillov2023segment}. To reduce overfitting to a particular prompt, ten different captioning instructions were manually generated. One of these instructions was then randomly selected to be associated with each image-description pair.

\vspace{-0.175cm}
\paragraph{$\bullet$ Stage II.} The second stage empowers Pheye to go beyond captioning, using a mixture of complex multi-modal instruction following examples: MIMIC-IT \citep{li2023mimic}, LRV \citep{liu2023aligning}, SViT \citep{zhao2023svit} and LVIS \citep{wang2023see}. This comes to a total of 964K samples. For each sample with multiple instruction-response turns, a single turn was randomly selected.

\vspace{-0.175cm}
\paragraph{$\bullet$ Stage III.} Previous stages used synthetic GPT-4 \citep{achiam2023gpt} and GPT-4V \citep{gpt4v} data for finetunning, which is naturally prone to noise and hallucinations. To alleviate this,  we further finetunned our models on a mixture of academic task-oriented VQA and captioning datasets, as well as some synthetic multimodal instruction following examples, based on LLaVA 1.5-mix-665K. We specifically considered ST-VQA \citep{biten2019scene}, TextVQA, and the LLaVAR finetunning dataset for better scene-text performance, removed the RefCOCO~\citep{refcoco} and VisualGenome~\citep{krishna2017visual} datasets, and sampled 2 random turns from the VQAv2 \citep{antol2015vqa} and GQA~\citep{hudson2019gqa} examples for each image. As a result, the model is only trained on $22.4\%$ of the VQAv2 and GQA examples in the original LLaVA 1.5 mix. We also sampled single turns from LLaVA, LLaVAR, OCR-VQA~\citep{ocrvqa} and OKVQA~\citep{marino2019ok}. From A-OKVQA~\citep{schwenk2022okvqa}, we randomly sampled unique image-question pairs. 


\paragraph{~}
\vspace{-0.05cm}

We trained the dense cross-attention layers, LoRA, positional embeddings, and LayerNorm layers, at every stage, using mixed precision Bfloat16. The frozen parameters, which correspond to the ViT and the language model, were loaded in Bfloat16. Resolutions were also kept constant across stages for the same model. The first stage uses a learning rate of 2e-4, the second stage uses 1e-4, and the third stage uses 5e-5. All stages use a cosine decay learning rate scheduler. We used gradient accumulation with an effective batch size of 128 across stages as well.

Since gradient accumulation for varying sequence lengths implicitly gives more weight to tokens in smaller sequences, we used sum reduction for the loss, instead of the typical mean reduction, tracking the number of tokens used to calculate the loss for a full batch. Before each optimization step, we divide the gradients by the number of output tokens, thereby mimicking a batch size of 128.

\subsection{Experimental Results}
\begingroup
\setlength{\tabcolsep}{10pt}
\begin{table*}[t!]
\centering
\small
\begin{tabular}{ccccccccc}
& & \multicolumn{3}{c}{VQAv2} & & \multicolumn{3}{c}{TextVQA} \\
\cline{3-5} \cline{7-9} \\[-1.5ex]
Model & Resolution & Bottom & Middle & Top & &  Bottom & Middle & Top \\ 
\hline & \\[-1.5ex] 
Pheye-x4 & 448 & 73.55 & 75.19 & 76.09 & &  40.67 & 44.23 & 52.74 \\ 
Pheye-x4 & 672 & 74.26 & 75.17 & 75.99 & &  46.83 & 48.43 & 52.48 \\ 
\hline & \\[-1.5ex]
\rowcolor{gray!25}
\%$\Delta$ & & +0.96 & -0.03 & -0.13 & &  +15.15 & +9.50 & -0.49 \\ 
\hline & \\[-1.5ex]
Pheye-x2 & 448 & 74.50 & 75.47 & 76.67 &  & 42.19 & 45.59 & 54.18 \\ 
Pheye-x2 & 672 & 74.73 & 76.25 & 76.71 & & 47.31 & 50.67 & 53.34 \\ 
\hline & \\[-1.5ex]
\rowcolor{gray!25}
\%$\Delta$ & & +0.31 & +1.03 & +0.05 & & +12.14 & +11.14 & -1.55 \\ 
\hline & \\[-1.5ex]
\end{tabular}
\caption{Relative change in VQA accuracy for VQAv2 and TextVQA instances, according to data tertiles that reflect the relative dimensions of relevant image areas.}
\label{tab:finer}
\end{table*}
\endgroup

Table~\ref{tab:main_results} summarizes our experimental results. With less training data and less trainable parameters, Pheye surpasses other generalist models of similar size in scence-text tasks, without using pre-extracted OCR tokens in the textual instructions. Note, for instance, that MobileVLM V2~\citep{chu2024mobilevlm} exhibits subpar performance on TextCaps, which suggests that the model relies heavily on the presence of OCR tokens in the instruction. 

The performance on scene-text tasks also increases more with a higher image resolution, compared to an increase in parameter count. The opposite happens for more general image understanding tasks, like VQAv2 and NoCaps. Increasing the resolution allows the model to capture finer details in images, while increasing the parameter count allows the model to learn more visual concepts, like objects and relationships between objects. 

The last row of Table~\ref{tab:main_results} presents a test in which the inputs to our best model are first down-scaled to 224x224, this way resulting in images without fine details. A higher performance drop is again seen on tasks such as TextVQA, confirming the importance of the high image resolution.

\subsection{Assessing the Use of Fine Image Details}
\label{finerdetails}

In order to further evaluate the impact that increasing the input image resolution has on tasks that involve the comprehension of fine-grained details, we followed a strategy similar to that of~\citet{zhang2023visual} and partitioned the TextVQA validation set into three groups of approximately equal size, based on the relative size of the ground-truth bounding box $S = \frac{A_{bb}}{A_{total}}$, where $A_{bb}$ denotes the area of the answer bounding box, and $A_{total}$ denotes the total area of the image. Specifically, we divided the data into three tertiles: the bottom tertile (finer details), the middle tertile (medium), and the top tertile (broader entities). The selection of the ground-truth bounding box was based on the average string similarity with all the ground truth answers, using the longest contiguous matching subsequence algorithm.

We also applied the same approach to a randomly sampled subset of the VQAv2 validation split, resulting in 10,000 questions pertaining to 8,453 images. For this dataset we used the segmentation area of objects as opposed to bounding boxes, since this quantity can better represent an object's area in the image, and used its category name for the string similarity algorithm. Since a large portion of VQAv2 answers correspond to yes/no or a number, in these cases we applied the same algorithm to the question, instead of the ground truth answers.

Table \ref{tab:finer} presents the results of our analysis. In both datasets, increasing resolution seems to lead to a greater improvement in accuracy for the samples that require understanding finer details within the image. This is particularly evident in TextVQA, as tertiles corresponding to smaller image-question-answer triplets have a greater relative improvement in performance, in comparison with larger tertiles. In VQAv2, however, this pattern is not as consistent, likely due to the less frequent appearance of category names for each object in questions and answers, compared to OCR tokens in TextVQA.

Appendix~\ref{sec:appendix2} further analyses how the model makes use of the high-resolution image inputs, presenting results on how the cross-attention module uses local versus global sub-images.

\section{Conclusions}

We presented an approach for building efficient Vision-Language Models (VLMs), processing high-resolution images while maintaining parameter efficiency. The approach was used to develop the Pheye family of VLMs, which achieve a high effectiveness on various academic task-oriented datasets, surpassing other generalist models of similar size, and achieving particularly strong results on tasks that involve understanding scene-text. 

Future work directions include investigating the use of different vision encoders, for instance incorporating strategies to process images at close to native resolutions and aspect ratios, e.g. building upon the approach proposed by~\citet{dehghani2024patch}. We can also explore ways to increase the amount of training data, given that our method still requires training hundreds of millions of randomly initialized parameters. To address this, we could generate additional task-specific synthetic data, particularly for tasks that involve scene-text, where the amount of available human generated data is smaller. Building on the work of~\citet{zhang2023visual}, we could generate VQA examples that focus on finer image details, which have been shown to be crucial for performance. 


\section*{Limitations and Ethical Considerations}

While our work does not raise new ethical issues within the domain of vision-language models (e.g., we conducted our experiments on public datasets, carefully designed for academic research and extensively used in previous studies), there are some general important concerns. 

Vision-Language Models (VLMs) are, for instance, notorious for their internal biases, inherited from the training data itself or from the use of pre-trained models such as CLIP. We therefore recommend caution in the use of the approach proposed in this paper, and anticipate further research into model biases, before relying on our work beyond research environments. 

Another important limitation in the work reported on this paper concerns the fact that our experiments relied exclusively on English datasets. Multilingual models have shown potential in leveraging diverse datasets and providing more robust and versatile language understanding capabilities, which could be beneficial for creating VLMs that can handle a wider variety of tasks and languages. Future work can perhaps explore the use of efficient multilingual models like Qwen~\citep{qwen} to enhance our approach, although additional efforts would be required in the design of an effective mixture of multilingual data for training.

\section*{Acknowledgements}

This research was supported by the Portuguese Recovery and Resilience Plan through project C645008882-00000055 (i.e., the Center For Responsible AI), and also by the Fundação para a Ciência e Tecnologia (FCT), specifically through the project with reference UIDB/50021/2020 (DOI: 10.54499/UIDB/50021/2020). 

\bibliography{custom}

\appendix

\section{Data Mixture for Final Training Stage}
\label{sec:datamix}

Table~\ref{tab:datamix} summarizes the datasets considered for Stage III of the proposed model training procedure, showing the response formatting prompt associated to each of the considered datasets.

\begin{table}[h!]
    \centering
    \tiny
    \begin{tabular}{p{0.39in} p{0.10in} p{2.1in}}
        Data & Size & Response Formatting Prompts\\
        \hline & \\[-1.5ex]
        LLaVA & 158K & - \\
        LLaVAR & 20K &  \\
        \hline & \\[-1.5ex]
        VQAv2 & 166K & Answer the question using a single word or phrase\\
        GQA & 144K & \\
        OK-VQA & 9K & \\
        OCR-VQA & 80K & \\
        TextVQA & 35K & \\
        ST-VQA & 26K & \\
        \hline & \\[-1.5ex]
        A-OKVQA & 17k & Answer with the option’s letter from the given choices directly. \\
        \hline & \\[-1.5ex]
        TextCaps & 22K & Generate a one-sentence caption for the provided image, incorporating textual elements visible in the image. \\
        \hline & \\[-1.5ex]
        Total & 676K & \\
        \hline & \\[-1.5ex]
    \end{tabular}
    \vspace{-0.25cm}    
    \caption{Data mixture used in Stage III of training.}
    \label{tab:datamix}
\end{table}

\section{Preliminary Experiments Assessing Architectural Choices}
\label{sec:appendix}

Through a set of preliminary experiments, we compared different approaches to combine the vision and language modalities, as well as approaches for encoding high resolution images, using relatively small Vision Transformers (ViTs) and language models in order to validate architectural decisions. 

The experiments followed the main principles of Flamingo~\citep{alayrac2022flamingo}, where the language and vision models were frozen to preserve their pre-training knowledge. 

\subsection{Vision-Language Model Architectures}

One initial experiment evaluated three different alternatives for combining the vision and language modalities, taking inspiration from the FROMAGe~\citep{koh2023grounding}, LLaVA, and SmallCap~\citep{ramos2023smallcap} neural architectures.

Building upon FROMAGe, our first approach involves transforming an image into a set of visual embeddings using a pre-trained ViT. The resulting visual inputs are then summarized through the extraction of the \texttt{[CLS]} token, which is subsequently mapped to the language model's dimensionality via a linear transformation. In contrast, the second architecture (i.e., the one inspired by LLaVA) leverages all the embeddings generated by the ViT to represent an image, rather than relying solely on the \texttt{[CLS]} token. In both cases, the language model takes as input the concatenation of the vision embeddings and the text prompt embeddings, allowing it to contextualize the image with the accompanying text, to generate a relevant response.

The third architecture, inspired by SmallCap, also makes use of a ViT that extracts features from an image. However, it differs from the last two architectures as it makes use of cross-attention modules inserted between the self-attention and the feedforward modules in the decoder, similarly to the original encoder-decoder Transformer, to bridge the modalities. The inner dimensionality of the cross-attention modules matches the hidden dimensionality of the language model.

\begin{table}[t!]
    \small
    \centering
    \begin{tabular}{lccccc}
                & Train & Param. & CIDEr & B@4 & M\\
        \hline  & \\[-1.5ex]
        FROMAGe & \ding{55} & 590K & 48.7 & 14.9 & 17.8\\
        FROMAGe & \ding{51} & 88M & 66.5 & 21.9 & 20.7\\
        \hline  & \\[-1.5ex]
        LLaVA & \ding{55} & 590K & 81.2 & 24.1 & 22.1 \\
        LLaVA & \ding{51} & 88M & 92.4 & 28.3 & 24.0\\
        \hline  & \\[-1.5ex]
        SmallCap & \ding{55} & 28M & 106.8 & 32.6 & 26.4 \\
        SmallCap & \ding{51} & 116M & 107.4 & 33.7 & 26.8\\
        \hline & \\[-1.5ex]
    \end{tabular}
    \caption{Comparison of different approaches for combining the vision and language modalities. The column "Train" denotes models in which the ViT was trained, and the column "Param." denotes the number of trainable parameters. "B@4" and "M" represent the BLEU-4 \citep{papineni-etal-2002-bleu} and METEOR \citep{banerjee-lavie-2005-meteor} metrics, respectively.}
    \label{preliminary1}
\end{table}

\begin{figure*}[t!]
        \centering
        \includegraphics[scale=.375]{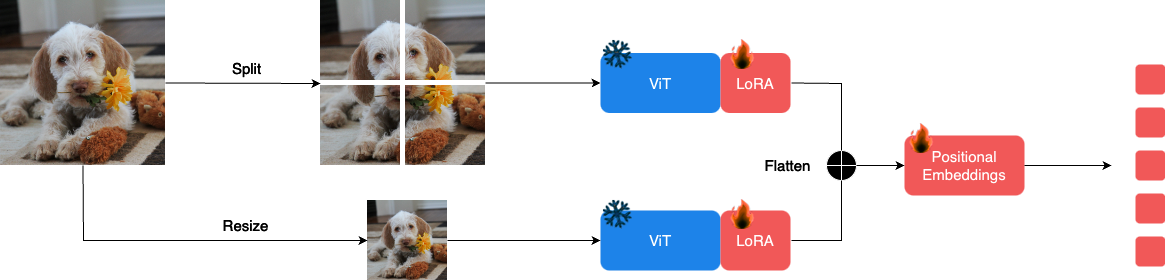}
        \caption{Architecture for high resolution multi-patch image encoding.}
        \label{fig:visionencoder}
\end{figure*}

To compare the effectiveness of the aforementioned three architectures, we trained and evaluated them on a captioning task using the COCO dataset~\citep{lin2014microsoft}. Specifically, we trained our models on the training split, using a cross-entropy loss, and evaluated them on the validation split. The ViT was initialized with pre-trained weights from a CLIP-ViT-B-32 model, while the language model was initialized with pre-trained weights from the GPT-2 base model~\citep{radford2019language}. During training, the language model was kept frozen, while the connector module, comprising either a linear mapping in the first two architectures or cross-attention modules in the third, was trained. Additionally, we experimented with training the ViT. The models were trained for 5 epochs using the AdamW optimizer, with a batch size of 64 instances, and a learning rate of 1e-4 for the cross-attention modules and 1e-5 for the vision encoder. The GPT-2 model was trained to generate captions using the prefix \texttt{"This image shows "}, and the model was also prompted with this prefix during the evaluation stage.

The results of this experiment are summarized in Table~\ref{preliminary1}, and they reveal that training the vision encoder together with the connector module yields improved performance across all architectures, when compared to training the connector module alone. This is expected, given that training the ViT results in training a larger number of parameters. Furthermore, our findings suggest that using the full image embeddings results in better performance relative to using solely the \texttt{[CLS]} token. This result is also intuitive, as multiple visual tokens can capture more nuanced details about the input image than a single vector. More importantly, in an experimental setup where the language model is not trained, the architecture inspired by SmallCap clearly demonstrates superior performance. Interestingly, using cross-attention modules as the modality bridge seems to outperform other architectures, like LLaVA, even when considering more trainable parameters associated to training the ViT.

Similar findings to those reported in this section have also been reported by~\citet{laurenccon2024matters}, although these authors have also showed that LLaVA-style linear projections can outperform cross-attention if, besides the ViT, the language model is also carefully trained.

\subsection{Encoding High Resolution Images}
\label{sec: vision encoder}

Previous attempts at increasing the input resolution of VLMs mostly chose to finetune the vision encoder on higher resolution images~\citep{chen2023pali}, which is costly due to the quadratic computational complexity of the attention mechanism, and which also requires large amounts of image-text data. To overcome this issue, we experimented with an approach similar to that of~\citet{li2023monkey}, in which we scale up the input resolution by splitting the image into smaller patches that match the resolution of the vision encoder, afterwards encoding the smaller sub-images individually. To provide the model with global context, we also encode the full image as normal, concatenate all feature maps, and use the result as our image representation.

Since the smaller image patches can have a different distribution than that of the images in which the ViT was trained on, we introduce two sets of LoRA adapters to the vision encoder. One is used on the global image and the other is used on the smaller local patches, allowing each resolution to specialize on different size aspects of the input image. The adapters were inserted at every linear layer of the ViT with the following hyperparameters: rank of 8, alpha of 16, and dropout probability of 0.05. We concatenate the resulting visual tokens, and add to each patch learned positional embeddings, as shown in Figure~\ref{fig:visionencoder}.

\begin{table*}[t!]
    \centering
    \small
    \begin{tabular}{ccccccc}
        Resolution & Resampler & Visual Token Length & Trainable Parameters & CIDEr & BLEU-4 & METEOR\\
        \hline & \\[-1.5ex]
        224 & - & 257 & 241M  & 104.5 & 27.6 & 24.6 \\
        448 & Monkey & 257 & 248M & 110.1 & 28.1 & 24.9 \\
        448 & Flamingo & 257 & 317M & 110.3 & 28.2 & 24.9 \\
        448 & - & 1285 & 244M & 113.2 & 28.5 & 25.2 \\
        \hline & \\[-1.5ex]
    \end{tabular}
    \caption{Comparison of different strategies for handling high resolution images as input.}
    \label{multipatcharch}
\end{table*}

The proposed method generates image representations with larger sequence lengths, which can be computationally demanding. To mitigate this issue, we explored the use of two resampler architectures, namely a 6-layer Perceiver Resampler, as introduced in Flamingo~\cite{alayrac2022flamingo}, and the Monkey Resampler, which can be characterized as a single layer Perceiver Resampler that does not include a feed-forward module and residual connections for the query vectors. In our experiments, both modules used 257 query vectors with the same dimensionality of the vision encoder. Additionally, we compare the aforementioned resamplers, which encode images at 448x448 pixels, with an encoder that processes images at 224x224 pixels, and also with a 448x448 pixels encoder that does not compress the image features. For a fair comparison, the smaller 224x224 pixels encoder also includes one set of LoRA adapters.

Due to the fact that scaling up the resolution is likely to benefit tasks that involve scene-text and OCR the most, we pre-train our models on the LLaVAR~\citep{zhang2023llavar} dataset for 1 epoch, finetune on the TextCaps~\citep{sidorov2020textcaps} train split for 3 epochs, and finally evaluate on the TextCaps validation split. When finetuning our models, we use the prompt \texttt{"This image shows "}, similarly to the previous experiment. Our vision Transformer and language model are initialized from pre-trained CLIP-ViT-L-14 and GPT-2 large models, respectively, and are combined with cross-attention modules, following SmallCap. The use of larger a ViT and language model, when compared to the experiment described in the previous appendix, relates to the fact that tasks involving scene-text and OCR are more demanding, and hence slightly larger models can facilitate the assessment of differences between the architectural alternatives being compared. We freeze the ViT and language model, and only train the cross-attention modules, positional embeddings, LoRA, and resamplers, using a cross-entropy loss over non-prompt tokens. We use the AdamW optimizer, with a learning rate of 1e-4 for the pre-training stage and 2e-5 for the fine-tuning stage, and with a batch size of 64 in both stages. The results are shown in Table~\ref{multipatcharch}.

The experiment revealed that increasing the input resolution from 224x224 to 448x448 pixels increases captioning performance, independently from whether we compress the resulting visual tokens or not. There does not seem to be a large performance difference in how we compress the visual tokens, since the Monkey Resampler and the Perceiver Resampler have similar CIDEr~\citep{vedantam2015cider} scores on the TextCaps validation split. However, this might not be the case for other tasks. Finally, the architecture with the best results is the one that increases the input resolution from 224x224 pixels to 448x448 pixels without compressing the vision features.

\section{Assessing Changes in Cross-Attention According to Different Path Sizes}
\label{sec:appendix2}

This appendix further analyses the use of high-resolution inputs, assessing the reliance of the model on the global versus local sub-images.

\begin{figure}[t!]
  \centering
  \includegraphics[width=0.48\textwidth]{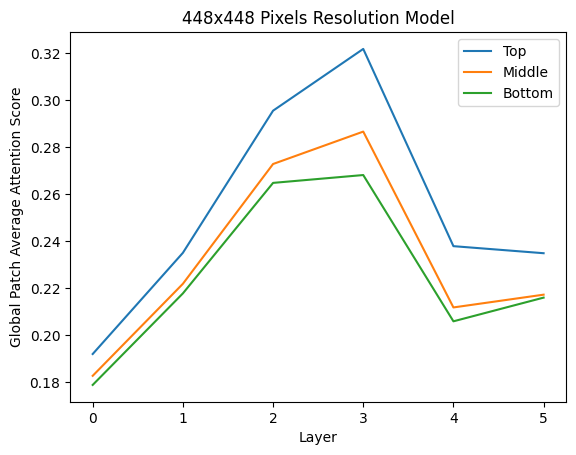}
  \includegraphics[width=0.48\textwidth]{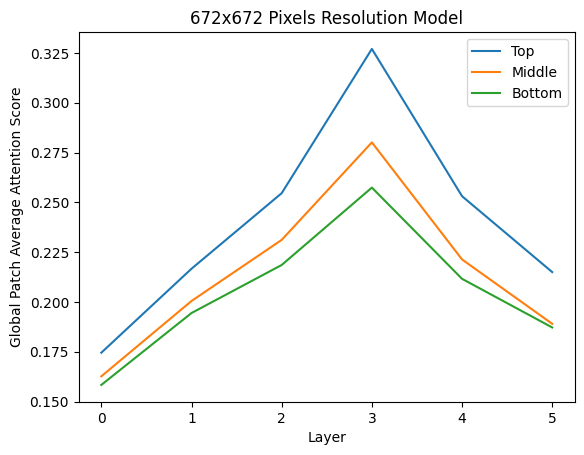}
  \caption{Attention scores for global patch tokens across data tertiles that reflect the relative dimensions of relevant image areas. Both graphs were calculated using the Pheye-x4 models. The average cross-attention score for the local patches is given by $1 - A_{G}$, were $A_{G}$ denotes the cross-attention scores for the global patch tokens.}
  \label{attnscores}
\end{figure}

To investigate how the model uses the different patches in the cross-attention module, we calculated the average of the attention scores for the generated captions across all TextCaps validation set images. Specifically we compute the average of the cross-modal attention scores across the global image tokens and the local patches tokens, at each step of generating caption tokens and across all attention heads. Figure \ref{attnscores} shows the global patch average attention scores for the Pheye-x4 models, separately for each of the layers in which cross-attention operations were inserted.

After computing the attention scores, we segmented the results using a similar approach to that described in Section~\ref{finerdetails}, with the only difference being that we replace the ground-truth answers with reference captions. This analysis reveals that images requiring finer detail understanding tend to favor local patches over global patch tokens, as they necessitate higher resolution inputs. In contrast, images with larger scene-text tend to rely more on global patch tokens. Moreover, we see that although the visual token length is different in both models present in Figure~\ref{attnscores}, the attention scores for the global patch are similar across all dense cross-attention layers.
\end{document}